\documentclass{llncs}
\usepackage{makeidx} 
\usepackage{multirow}
\usepackage{color}
\usepackage{graphicx}
\usepackage{diagbox}
\usepackage{array}

\begin{document}

\title{Deep Learning Framework for Multi-class Breast Cancer Histology Image Classification} 

\author{Yeeleng S. Vang \and Zhen Chen\and Xiaohui Xie}
\institute{University of California Irvine\\ Irvine, CA 92697\\ \email{\{ysvang, zhenc4\}@uci.edu}  \email{xhx@ics.uci.edu}}

\maketitle

\begin{abstract}
In this work, we present a deep learning framework for multi-class breast cancer image classification as our submission to the International Conference on Image Analysis and Recognition (ICIAR) 2018 Grand Challenge on BreAst Cancer Histology images (BACH).  As these histology images are too large to fit into GPU memory, we first propose using Inception V3 to perform patch level classification.  The patch level predictions are then passed through an ensemble fusion framework involving majority voting, gradient boosting machine (GBM), and logistic regression to obtain the image level prediction. We improve the sensitivity of the Normal and Benign predicted classes by designing a Dual Path Network (DPN) to be used as a feature extractor where these extracted features are further sent to a second layer of ensemble prediction fusion using GBM, logistic regression, and support vector machine (SVM) to refine predictions. Experimental results demonstrate our framework shows a 12.5$\%$ improvement over the state-of-the-art model.

\end{abstract}

\section{Introduction}
In the United States, breast cancer continues to be the leading cause of cancer death among women of all races \cite{breast_cancer_stats}.  Studies have shown that improvement to survival rate over the last decade can be attributed to early diagnosis and awareness of better treatment options \cite{saadatmand2015},\cite{berry2005},\cite{gelder2015}. Common non-invasive screening test includes clinical breast exam which involves a visual check of the skin and tissue and a manual check for unusual texture or lump, mammography which requires taking an x-ray image of the breast to look for changes, and breast MRI which uses radio waves to obtain a detailed image inside the breast. Of the latter two diagnostic modals, many computer-aided diagnosis (CAD) systems have been developed to assist radiologists in their effort to identify breast cancer in its early stages \cite{hadjiiski2006}. On the other side of the screening toolbox are biopsies which are minimally invasive procedures whereby tissue samples are physically removed to be stained with hematoxylin and eosin (H\&E) and visualized under a microscope. These histopathology slides allow pathologists to distinguish between normal, non-malignant, and malignant lesions \cite{araujo2017} to assist in their diagnosis.  However, even among trained pathologists the concordance between their unanimous agreement is a mere 75\% \cite{elmore2015}.  This high degree of discord motivates the development of automatic CAD systems using machine learning to assist these professionals in their diagnosis.

From November 2017 to January 2018, the International Conference on Image Analysis and Recognition (ICIAR) held the 2018 Grand Challenge on BreAst Cancer Histology images (BACH) to solicit submissions of automatic image analysis systems for the task of four-class classification of breast cancer histology images.  Here we present a deep learning framework for the task of multi-class breast cancer histology image classification.  Our approach uses the Inception (GoogLeNet) V3 \cite{szegedy2015} architecture to discriminate between invasive carcinoma, in situ carcinoma, benign lesion, and normal tissue patches.  We then fuse these patch level predictions to obtain image level prediction using an ensemble framework.  Our system improves the sensitivity over the benign and normal classes by using a Dual Path Network (DPN) \cite{chen2017} to extract features as input into a second level ensemble framework involving GBM, SVM, and logistic regression. Experimental results on a held out set demonstrate our framework shows a 12.5$\%$ improvement over the state-of-the-art model.

\section{Relate Work}
Several works have been published in the area of applying machine learning algorithms for cancer histology image detection and classification \cite{nayak2013},\cite{gorelick2013},\cite{xu2014},\cite{wang2010}. In the specific area of breast cancer histopathology classification, the Camelyon 16 competition led to numerous new approaches utilizing techniques from deep learning to obtain results comparable to highly trained medical doctors \cite{harvard},\cite{google},\cite{babak2017}. The winning team used Inception V3 to create a tumor probability heatmap and perform geometrical and morphological feature selection over these heatmaps as input into a random forest classifier to achieve near 100$\%$ area under the receiver operating characteristic curve (AUC) score \cite{harvard}. However, this competition involved only binary class prediction of tumor and normal whole slide images.  For 4-class breast cancer classification, Araujo et al. \cite{araujo2017} published a bespoke convolutional neural network architecture that achieved state-of-the-art accuracy results and high sensitivity for carcinoma detection.  

\section{ICIAR$2018$ Grand Challenge Datasets and Evaluation Metric}
In this section, we describe the ICIAR$2018$ dataset provided by the organizers for the subchallenge of multi-class breast cancer histology image classification and the evaluation metric used to score submissions.  The interested reader is encouraged to refer to the competition page for details regarding the other subchallenge.

\subsection{ICIAR$2018$ Dataset}
The ICIAR2018 breast cancer histology image classification subchallenge consist of Hematoxylin and eosin (H$\&$E) stained microscopy images as shown in Table \ref{dataset}. The dataset is an extended version of the one used by Araujo et al. \cite{araujo2017}. All images were digitized with the same acquisition conditions, with resolution of $2040 \times 1536$ pixels and pixel size of $0.42\mu m \times 0.42\mu m$. Each image is labeled with one of four classes: i) normal tissue, ii) benign lesion, iii) in situ carcinoma and iv) invasive carcinoma according to the predominant cancer type in each image. The images were labeled by two pathologists who only provided a diagnostic from the image contents without specifying the area of interest. There are a total of $400$ microscopy images with an even distribution over the four classes. We randomly perform a $70\%$-$20\%$-$10\%$  training-validation-test split.  The training and validation sets are used for model development while the test set is held out and only used for evaluation.

\begin{table}
\centering
\caption{ICIAR$2018$ H\&E Histopathology Dataset}
\begin{tabular}{c|c|c|c|c|c} 
\hline
\multicolumn{2}{c|}{Type}                     & Training & Validation & Test  & Total \\ \hline
\multirow{4}{*}{Microscopy} & normal   & $70$   &       $20$      & $10$ &\multirow{4}{*}{$400$} \\ 
                                            & benign   & $70$   &       $20$      & $10$ & \\ 
                                            & in situ     & $70$   &       $20$      & $10$ & \\ 
                                            & invasive & $70$   &       $20$      & $10$ & \\ \hline

\end{tabular} \label{dataset}
\end{table}

\subsection{Evaluation Metric}
This challenge consists of automatically classifying H$\&$E-stained breast cancer histology images into four classes: normal, benign, in situ carcinoma and invasive carcinoma.  Performance on this challenge is evaluated based on the overall prediction accuracy, i.e. the ratio of correct predictions over total number of images.

\section{Method}
In this section, we describe our framework and approach to this problem of multi-class breast cancer histology image classification.

\subsection{Image-wise classification of microscopy images}
\subsubsection*{Stain Normalization Pre-processing}
Stain normalization is a critically important step in the pre-processing of H\&E stain images. It is known that cell nucleus are stained with a large amount of pure hematoxylin and a small amount of Eosin whereas cytoplasm is stained with a large amount of pure eosin and small amount of hematoxylin \cite{vahadane2015}.  Variations in H\&E images can be attributed to such factors as differences in lab protocols, concentration, source manufacturer, scanners, and even staining time \cite{babak2016}. These variations makes it difficult for software trained on a particular stain appearance \cite{khan2014} therefore necessitates careful preprocessing to reduce such variances.

Many methods have been proposed for stain normalization including \cite{khan2014},\cite{macenko2009},\cite{vahadane2015} that are based on color devolution where RGB pixel values are decomposed into their stain-specific basis vectors. In addition to color information, Bejnordi et el. takes advantage of spatial information to perform this deconvolution step \cite{babak2016}, however their approach currently only works for whole slide images.

\begin{figure}[]
   \centering
   \includegraphics[width=1\textwidth]{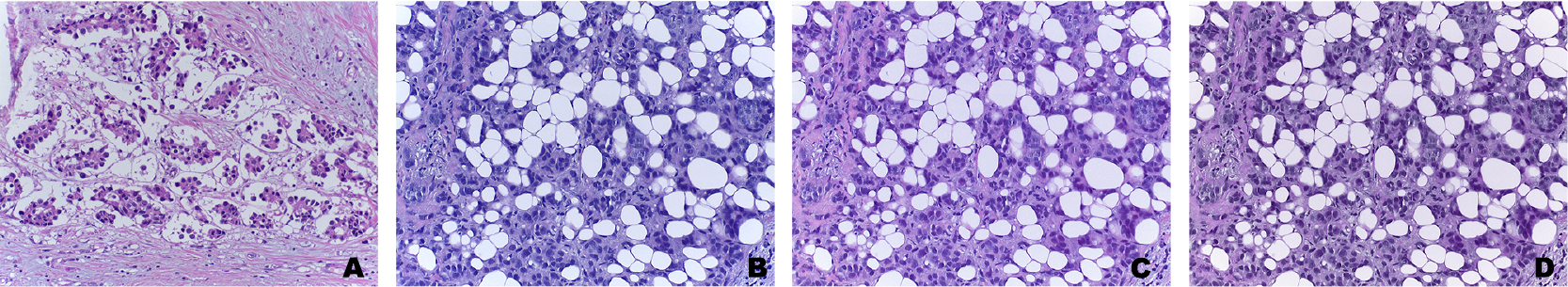}
   \caption{$A$ target image, $B$ original image, $C$ image after Macenko normalization, $D$ image after Vahadane normaliztion.}
   \label{fig}
\end{figure}

In our framework, we utilized both Macenko \cite{macenko2009}, which used singular value decomposition (SVD), and Vahadane normalizations \cite{vahadane2015}, which used sparse non-negative matrix factorization (SNMF), as part of our ensemble framework.  This was due to the fact that initial empirical results showed Macenko-normalized images obtained high sensitivity for invasive and in situ classes whereas Vahadane-normalized images showed high sensitivity for benign and normal classes.  Both set of normalized datasets were normalized using "iv001.tif" as the target image. An example of both normalization schemes are shown in Fig. \ref{fig}.

\subsubsection{Image-wise Classification Framework}
\begin{figure}
   \centering
   \includegraphics[width=1\textwidth]{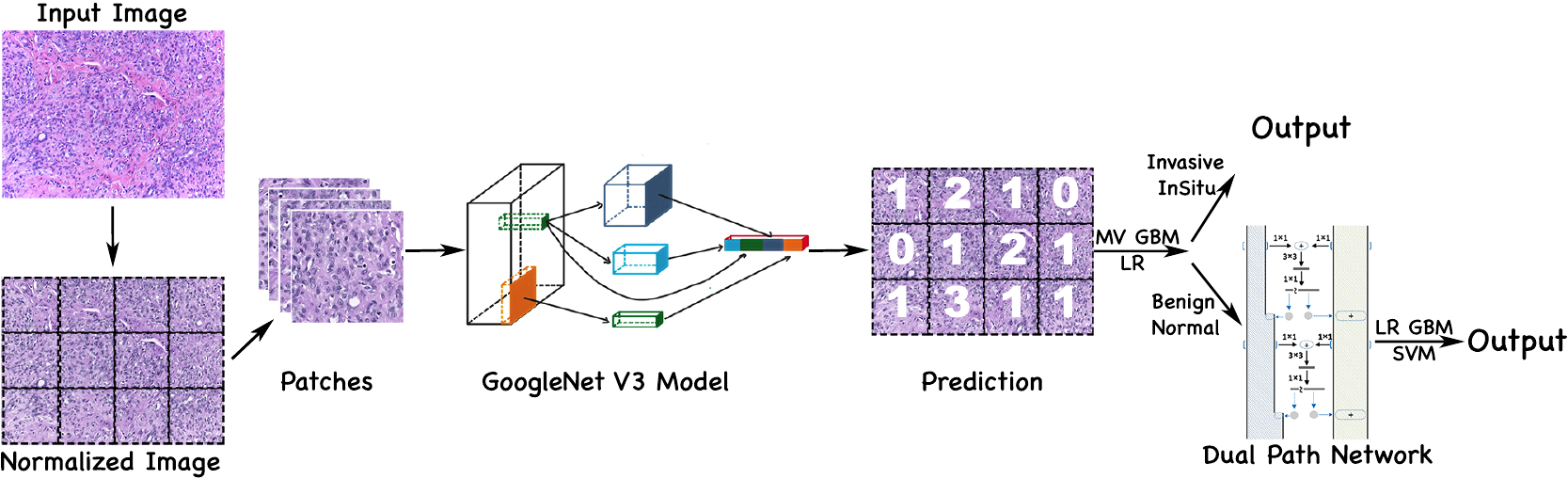}
   \caption{The framework of image-wise classification. The normalized input image is patched into twelve non-overlapping patches.  8 sets of these patches are generated corresponding to the 8 valid orientations.  These 8 sets of patches are passed through the Inception (GoogLeNet) V3 model to generate a patch level heatmap probability tensor.  The heatmap tensor is then fused using majority voting (MV), gradient boosting machine (GBM), and logistic regression (LR) across both macenko-normalized and vahadane-normalized version of the input image. If the model predicts invasive or in situ carcinoma, the model outputs this prediction.  Otherwise the normalize images are pass through the DPN network to extract features for a second fusing step involving LR, GBM, and support vector machine (SVM) to output prediction for benign and normal class. }
   \label{framework}
\end{figure}

The microscopy classification framework consists of a patch level classification stage, an image level heatmap post-processing stage, and possibly a refinement stage, as depicted in Fig. \ref{framework}. During model training of the patch-based classifer, each patch input is of size $512\times512$. We extracted $500$ patches from each microscopy slide in the training and validation sets for both Macenko-normalized and Vahadane-normalized datasets. $35$ of those patches comes from sliding over the  normalized microscopy image with strides of $256$ while the remaining patches were randomly sub-sampled. As with the assumption used in \cite{araujo2017}, these patches are given the same label as the original slide image with which they where obtained from. 

A pretrained Inception V3 model \cite{szegedy2015}, is modified to accept image patch of this size and trained to discriminate between the four classes.  At training time, images data are dynamically augmented before being fed through the model. Similar to the color perturbation scheme used in \cite{google}, brightness is perturbed with a delta of $5/255$, contrast with a delta of $.05$,  saturation with a delta of $.05$, and hue with a delta of $0.02$. In addition to color perturbation, images were randomly flipped vertically and/or horizontally, and randomly rotated by 90 degrees to obtain all eight valid orientations. 

The Inception V3 model was fine-tuned on 4 GPUs (2 Nvidia Titan X GPUs and 2 Nvidia GTX 1080Ti) where each GPUs receive a batch of 8 images.  Model is trained for 30 epochs with learning rates set as: 5e-5 for the bottom 5 convolution layers, 5e-4 for the eleven inception modules, and 5e-2 for the top fully connected layer. Learning rate was decreased by 0.95 every 2 epochs. The RMSprop optimizer \cite{rmsprop} with 0.9 momentum is used and the best performing model on the validation set is saved.

At inference time for a single microscopy image, a heatmap tensor of size [$8\times4\times3\times4$] is obtained. The first dimension corresponds to the $8$ valid orientations of the image, the second dimension to the $4$ classes, and the third and fourth dimension corresponds to the spatial dimensions of the image using non-overlapping patching.  

\subsubsection{Heatmap-based Post-processing}

Three data fusion strategies were investigated for this competition.  The first strategy involved finding the average probabilities along the first dimension of the heatmap and then assigning labels to each $3\times4$ patches corresponding to the most probable class, which we will call the class map.  From this $3\times4$ class map, a final label for the microscopy is obtained by majority voting. The second and third strategies involved finding the class map for each of the $8$ orientation separately first, and then obtaining a histogram of the classes across all $8$ orientations.  The histogram data is then used to train two separate models: a logistic regression with $L1$ regularization and a gradient boosting machine (GBM) classifier (num. of estimator=280, max depth = 4, learning rate = .9) to ultimately classify the image similar to \cite{hou2016}.  If the model predicts benign or normal, the vahadane-normalized image was further passed through a refinement stage as will be describe in the next section.

\subsubsection{Refinement model for Benign and Normal classes}

Since the Inception model yielded low sensitivity for both normal and benign classes with many interclass misclassification between these two classes, we proposed training a slimed-down version of the dual path network (DPN) \cite{chen2017} to serve as a feature extractor for use with Vahadane-normalized images.  DPN was chosen due to its compact size and having beneficial characteristics of both residual-like and densenet-like architectures.  Using the features extracted by the DPN, we train three additional models: GBM, Support Vector Machine (SVM), and Logistic regression with $L1$ for binary classification. The results for our entire pipeline is presented below in Table \ref{partA}.

\section{Experimental Results}

The performance of our framework on image-wise classification is shown below in Table \ref{partA}.  As a baseline, we compare against Araujo et al. \cite{araujo2017} which, although using a smaller subset of this dataset, tested on a held-out set of roughly the same size. Their best accuracy performance on this 4-class classification problem was 77.8$\%$.  Our framework achieves an accuracy score of 87.5$\%$, a 12.5$\%$ improvement over the baseline score. Even without the refinement model, our model offers a 6 $\%$ improvement over the baseline.

\begin{table}
\centering
\caption{Image-wise Classification Results}
\begin{tabular}{c|c|>{\centering\arraybackslash}p{2.5cm}|>{\centering\arraybackslash}p{2.5cm}} 
\hline
\multicolumn{2}{c|}{}     &  \multicolumn{2}{c}{Accuracy} \\ \cline{3-4}    
\multicolumn{2}{c|}{}     &  Validation Set  & Test Set       \\ \hline
\multirow{3}{*}{\begin{tabular}{@{}c@{}}Macenko \\ normalization\end{tabular}}      & MV       &  $0.800$ & $0.775$ \\  
                                                                                                                                    & LR &  $0.750$ & $0.775$ \\                           
                                                                                                                                    & GBM    &  $0.775$ & $0.775$ \\ 
\hline 
\multirow{3}{*}{\begin{tabular}{@{}c@{}}Vahadane \\ normalization\end{tabular}}      & MV         &  $0.788$ & $0.775$ \\  
                                                                                                                                    & LR &  $0.763$ & $0.775$ \\                           
                                                                                                                                    & GBM   &  $0.750$ & $0.800$ \\ 
\hline                                                                                                                                     
\multicolumn{2}{c|}{Ensemble}                                                                                                                        & $0.825$  & $0.825$ \\                                                                                                                                 
\hline                                                                                                                                     
\multicolumn{2}{c|}{Ensemble with refinement}                                                                                                                        & $0.838$  & $0.875$ \\                                                                                                                                 
\hline
\end{tabular} \label{partA}
\end{table}

Comparing the sensitivity by Araujo et al. \cite{araujo2017}, we see they achieved sensitivities of 77.8 $\%$, 66.7$\%$, 88.9$\%$, and 88.9$\%$ for normal, benign, in situ, and invasive classes respectively.  From Table \ref{contingency}, we showed higher sensitivity across all four classes using our framework. Of noticeable improvement is the benign class which we saw an almost 20$\%$ improvement. This validates our decision to incorporate a binary class refinement phase specifically for the benign and normal classes.

\begin{table}
\centering
\caption{Image-wise Test Set Contingency Table}
\begin{tabular}{c|c|c|c|c|c} 
\hline
\backslashbox{\tiny{Ground Truth}}{\tiny{Prediction}}     & invasive & in situ & benign & normal & sensitivity \\ \hline
invasive & $9$        & $0$    & $1$      & $0$        & $0.90$ \\ 
in situ     & $0$        & $10$  & $0$      & $0$       & $1.00$ \\ 
benign   & $1$        & $1$    & $8$      & $0$        & $0.80$ \\ 
normal   & $0$        & $0$    & $2$      & $8$     & $0.80$ \\ 
\hline
\end{tabular} \label{contingency}
\end{table}

\section{Discussion}

In this work we proposed a deep learning framework for the problem of multi-class breast cancer histology image classification.  To leverage the advances from the computer vision field, we propose using the successful inception V3 model for initial four-class classification.  We propose a new ensemble scheme to fuse patch probabilities for image-wise classification.  To improve the sensitivity of the benign and normal class, we propose a two-class refinement stage using a dual path network to first extract features from the vahadane-normalized images and then using gradient boosting machine, support vector machine, and logistic regression to fuse all our predictions into a final result.  Experimental results on the ICIAR2018 Grand Challenge dataset demonstrates an improvement of 12.5$\%$ over the state-of-the-art system.


\end{document}